 \documentclass[sigconf, nonacm, screen]{acmart}
\AtBeginDocument{%
  \providecommand\BibTeX{{%
    \normalfont B\kern-0.5em{\scshape i\kern-0.25em b}\kern-0.8em\TeX}}}

\setcopyright{acmcopyright}
\copyrightyear{2018}
\acmYear{2018}
\acmDOI{10.1145/1122445.1122456}

\usepackage{algorithm}
\usepackage{algorithmic}
\usepackage{flushend}
\usepackage{lipsum}
\usepackage{float}
\usepackage{perpage}
\usepackage{color}
\usepackage{tablefootnote}
\usepackage{multicol}
\usepackage{flushend}
\usepackage{graphicx}
\acmConference[Woodstock '18]{Woodstock '18: ACM Symposium on Neural
  Gaze Detection}{June 03--05, 2018}{Woodstock, NY}
\acmBooktitle{Woodstock '18: ACM Symposium on Neural Gaze Detection,
  June 03--05, 2018, Woodstock, NY}
\acmPrice{15.00}
\acmISBN{978-1-4503-XXXX-X/18/06}

\usepackage[utf8]{inputenc}
\usepackage[T1]{fontenc}
\begin{document}

\title{An automated approach to mitigate transcription errors in braille texts for the Portuguese language}

\author{André Roberto Ortoncelli}
\affiliation{%
  \institution{Federal University of Technology - Paraná - (UTFPR)}
  \city{Dois Vizinhos}
  \state{Paraná}
  \country{Brazil}
}
\email{ortoncelli@utfpr.edu.br}
\author{Marlon Marcon}
\affiliation{%
  \institution{Federal University of Technology - Paraná - (UTFPR)}
  \city{Dois Vizinhos}
  \state{Paraná}
  \country{Brazil}
}
\email{marlonmarcon@utfpr.edu.br}
\author{Franciele Beal}
\affiliation{%
  \institution{Federal University of Technology - Paraná - (UTFPR)}
  \city{Dois Vizinhos}
  \state{Paraná}
  \country{Brazil}
}
\email{fbeal@utfpr.edu.br}

\renewcommand{\shortauthors}{Ortoncelli, Marcon and Beal} 

\begin{abstract}

The quota system in Brazil made it possible to include blind students in higher education. Teachers' lack of knowledge about the braille system can represent a barrier between them and students who use it for writing and reading. Computer-vision-based transcription solutions represent mechanisms for reducing understanding restrictions on this system. However, such tools face nuisances inherent to image processing systems, e.g., illumination, noise, and scale, harming the result. This paper presents an automated approach to mitigate transcription errors in braille texts for the Portuguese language. We propose a selection function, combined with dictionaries, that provides the best correspondence of words based on their braille representation. We validated our proposal on a dataset of synthetic images by submitting them to different noise levels and testing the proposal's robustness. Experimental results confirm the effectiveness of the solution compared to a standard approach. As a contribution of this paper, we expect to provide a method to support robust and adaptable solutions to real use conditions.

\end{abstract}




\maketitle

\section{Introduction}


The conception of new technologies to assist blind people is a critical challenge of the research community. The World Health Organization (WHO) estimates over 285 million blind and visually impaired people globally, whose 39 million are blind \citep{Who2020}. According to the report ``The Conditions of Eye Health in Brazil 2019,'' prepared by the Brazilian Council of Ophthalmology (CBO), and based on data from WHO reports and indexes from the Brazilian Institute of Geography and Statistics (IBGE), estimates that in Brazil, 1.577,016 individuals are blind, equivalent to 0.75\% of the population \citep{Almeida2019}. 

The quota system in Brazil allowed blind students to have access to the University. Despite the social advance in recent years, the current teaching model brings many difficulties to realize these students' inclusion. One of these difficulties relies on written communication between teacher and student, typically made using braille, a tactile reading and writing system.

\begin{figure}[htb]
  \centering
  \includegraphics[width=2cm]{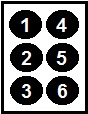}
  \caption{A full braille cell. Each dot on the cell must have a corresponding position on the braille map code, starting from the top-left position following the numerical order.}
  \Description{A full braille cell with six raised dots distributed in two parallel columns with three dots, left column points 1-2-3 and right column points 4-5-6.}
  \label{fig:cela_braille}
\end{figure}

\begin{figure*}[htb]
  \centering
  \includegraphics[width=0.8\linewidth]{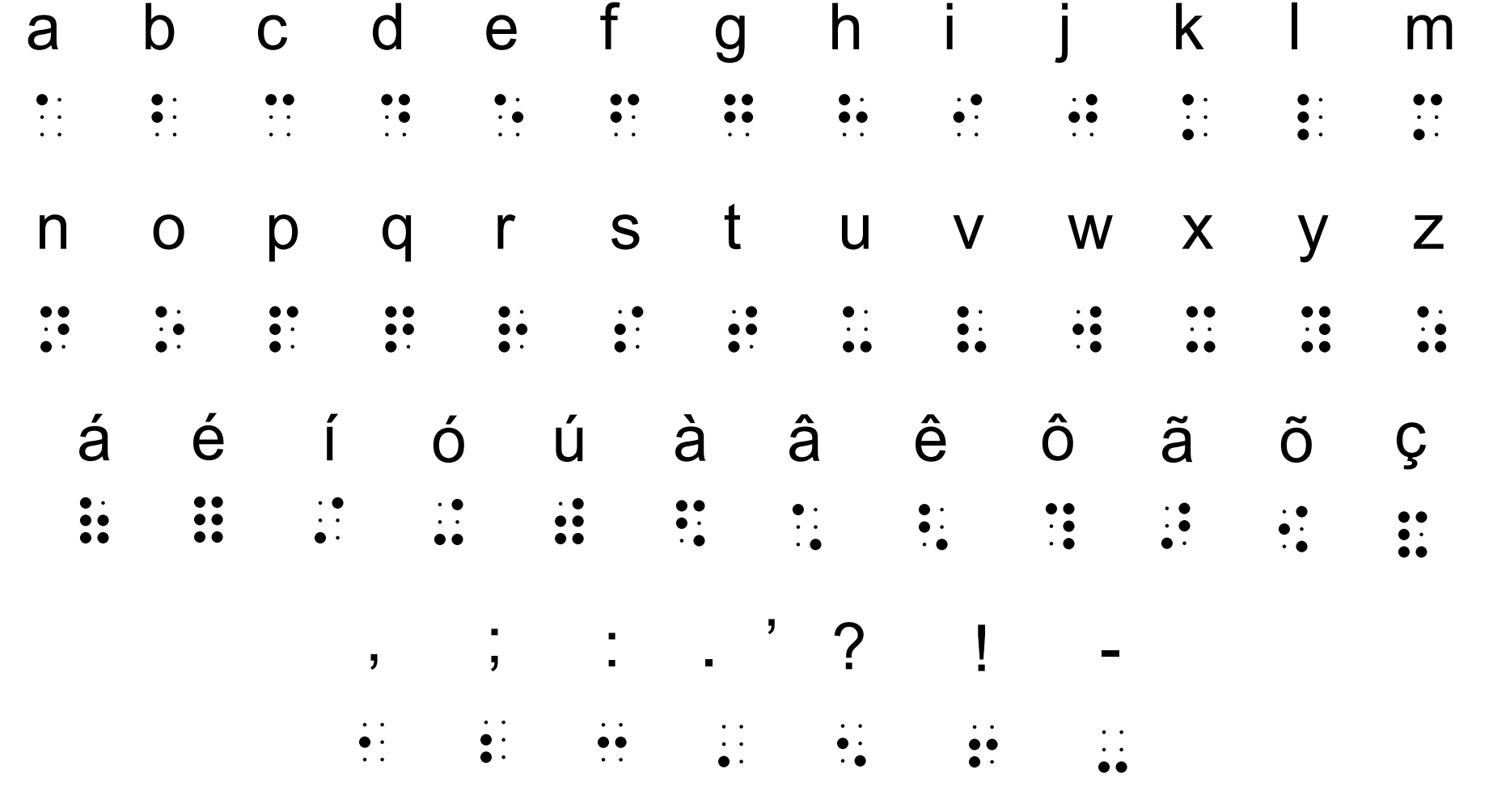}
  \caption{Examples of letters and symbols of the braille alphabet for the Portuguese language.}
  \label{fig:alfabeto-braille}
\end{figure*}

The braille system \citep{AFBlind2020}, proposed by Louis Braille in the middle of the 19th century, is a universal writing and reading system. Each character representation uses a rectangular cell of palpable dots, arranged in a 3 x 2 fashion, and presented in \autoref{fig:cela_braille}. To express letters, numbers, punctuation marks, and other symbols are defined as a combination of 6 dots, i.e., 64 combinations. In \autoref{fig:alfabeto-braille}, we depict some examples of the braille alphabet for the Portuguese language.

The vast majority of teachers do not know how to read braille and face comprehension problems, which difficulties evaluating their students' activities. To improve communication and interaction between the teacher and the blind student is a crucial feature of the teacher-student relationship. 

Computer-vision-based solutions can help to recognize students' texts reducing such restrictions. In the literature, we find many approaches to deal with the optical braille recognition (OBR) task \cite{Isayed2015}, i.e., transcribe a braille text into an alphanumeric representation \cite{antonacopoulos2004robust, stanco2013automatic,li2020optical}. These solutions must handle two significant problems on the images: the preprocessing and the segmentation stages. The first stage deals with inherent nuisances on the images, e.g., color, luminance, and contrast variations, as well as with noise, scale, and rotation variations that harm the recognition process. The second stage segments each braille cell. Both are complementary stages, and the first implies strongly on the second. 

As an indefectible process is almost inconceivable, some transcription errors eventually occur, especially when dealing with true images. Dictionaries can help with such transcription errors. However, spell-checking solutions use probabilistic aspects on a recommendation, that are strictly related to orthographic issues \cite{pyspellcheker}. 

This work proposes an automated approach to mitigate transcription errors in braille texts for the Portuguese language. We claim that using a braille-cell-oriented solution can provide a robust tool to OBR systems, principally related to Computer Vision (CV) recognition errors. To the best of our knowledge, this is the first braille-centered approach in the literature.


\begin{figure}[htb]
  \centering
  \includegraphics[width=0.85\linewidth]{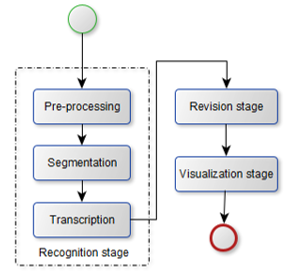}
  \caption{Flowchart of our proposed pipeline. Given a braille document image as input, our proposal preprocesses, segments, and transcribes braille cells into alphanumeric characters. Then we perform a braille-centered revision stage, and finally, the output text can be used for visualization.}
  \label{fig:process}
\end{figure}

Our proposed pipeline for transcribing braille image texts into plain text, and suggesting likely words to mitigate recognition errors, has three stages: the recognition, the revision, and visualization stages. Given an image of a braille text, our pipeline converts braille cells into characters by preprocessing the image, then segmenting into lines and letters, and finally converting them into alphanumeric values, following the braille code (e.g., \autoref{fig:alfabeto-braille}). The recognition stage outputs a transcribed text that serves as input to the revision stage, verifying the transcription result in the dictionary and proposing a correction when required. The visualization stage is conceptual, thinking in an application scenario, and in this work, only saves the transcription results in a text file. We depict this process in \autoref{fig:process}. 




\section{Proposed approach}
\label{sec:approach}

In this section we detail our proposed apprach in depth. In Section \ref{subsec:reconhecimento} reports the computer vision method employed on the recognition stage, and Section \ref{subsec:algoritmo} describer our proposed approach for revising recognition errors in braille texts. The source code of our porposed pipeline is available on \url{http://github.com/ICDI/braille-spellchecker}.

\subsection{Recognition stage}
\label{subsec:reconhecimento}

In this work, the transcribing process of a braille text into the Portuguese language has three steps: preprocessing, lines and columns (characters) segmentation, and conversion of the image representing the braille cell to the corresponding alphanumeric symbol\footnote{Our code enhances the code provided in \url{https://github.com/MUSoC/Braille-OCR}, involving the adaptation to deal with our dataset's images}.


\begin{figure*}[htb]
  \centering
  \includegraphics[width=1\linewidth]{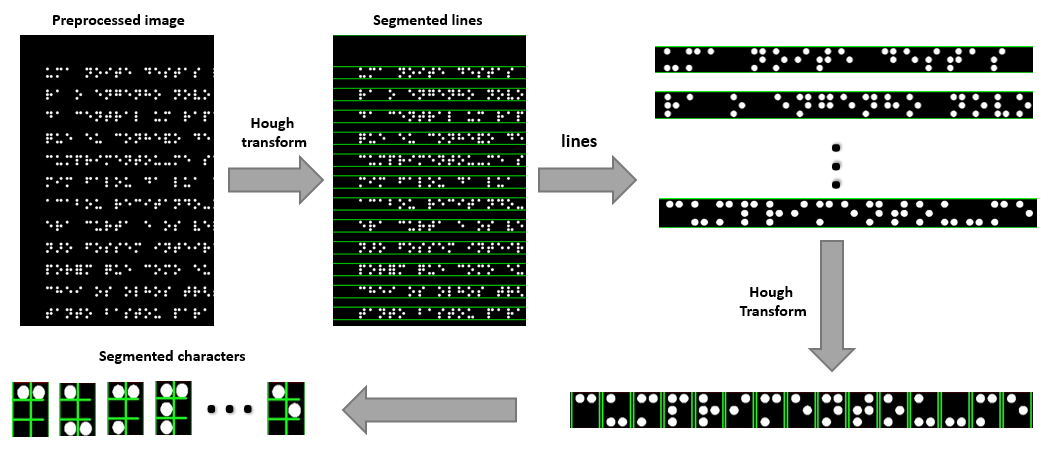}
  \caption{Segmentation pipeline. The algorithm segments page line by starting from a preprocessed image by extracting horizontal Hough segments and cropping the image. After, we perform a similar approach to segment chars by detecting vertical Hough lines. Finally, we split each character segment into points to transcribe the correspondent alphanumeric value. }
  \label{fig:char-segmentation}
\end{figure*}

\subsubsection{Preprocessing}

this step prepares the input image for the segmentation process. In this work, as we adopted synthetic images, preprocess the images is relatively more straightforward. However, in circumstances that demand the recognition of images captured from printed texts, this step is fundamental for the application's success. It requires more refined and adaptable techniques to the problem. We first remove noise from the images, using the following image processing techniques: 1) Image thresholding with a value fixed at 120, found empirically; 2) Median filter with $5 \times 5$ filter; 3) Morphological opening filter with a squared structuring element of $3 \times 3$; 4) Blob removal (area $ <10 $).


\subsubsection{Segmentation}

in this process, we first segment horizontally (rows) and then vertically (columns). Both follow the same principle, and we show the complete procedure in \autoref{fig:char-segmentation}. Starting from the image resulting from the pre-processing, initially, we apply a morphological expansion filter with a structuring element of $ 10 \ times l $, where $ l = $ page width. After that, we perform a Canny filter for edge detection, and finally, we detect the lines that horizontally segment the images, using the Hough line transform. After segmenting rows, we perform a similar process for column segmentation, i.e., the braille characters. Finally, we divide each image portion corresponding to a braille cell into six regions that correspond to the points. We convert each detected dot to a position on the six-position binary vector.


\subsection{Revision stage}
\label{subsec:algoritmo}

To understand our proposed revision method is essential to understand some assumptions about the domain of our work. Given the word denoted by a set of letters $P=[l_1, l_2, ..., l_x]$, each letter is part of a pre-defined set of characters. A text is a set of words denoted by $T=[p_1, p_2, ..., p_n]$. Like a text, a dictionary is also a set of words denoted by $D=[p_1, p_2, ..., p_m]$. 

In this work context, two aspects separate a text from a dictionary: i) a text can have repeated words. A dictionary has only one instance of each term; ii) in a dictionary, every element is grammatically correct. We use the words in a dictionary as a parameter to correct possible errors in a text.

\begin{figure}[htb]
  \centering
  \includegraphics[width=0.95\linewidth]{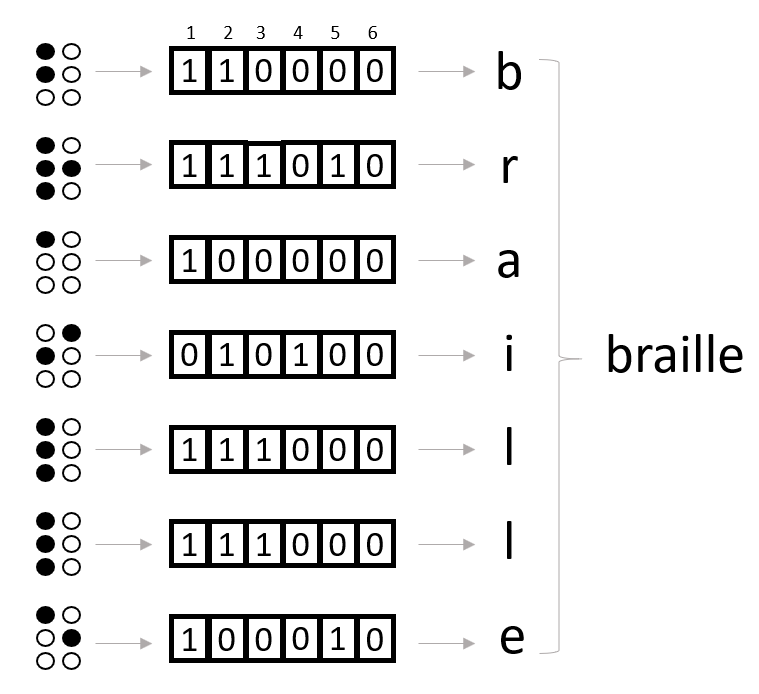}
  \caption{Transcription example of the ``braille'' word. Each braille cell has six dots that represent a binary code of the same length. We then map each code list into an alphanumeric character.}
  \label{fig:trascricao}
\end{figure}

We can transcribe each letter of a word $l_x \in P$ into the Braille system. The function $\beta(P)$ takes a letter $P$ and returns a vector of such size $| P |*6$. Each position is a binary value that corresponds to each of the six points of a braille cell. Similarly, the $\psi(\beta(P))$ function receives a braille cells vector and returns the word transcribed in the target language. In \autoref{fig:trascricao} we present an example of the transcription pipeline of a set of braille cells into a word. 

The proposed method takes a text ($ T $) and dictionary ($ D $) and uses $ D $ as a reference to correct possible grammatical errors in the words of $ T $.

The algorithm uses the function $ compare (p_t, p_d) $ to compare a word $ p_t \in T $ with a word $ p_d \in D $ (where $ | p_d | = | p_t | $). Equation \ref{eq:compare} defines the value returned by the $compare$ function.

 \begin{equation}
 \label{eq:compare}
 compare(p_t, p_d)  =  \sum_{i=1}^{|p_t|*6} |\beta(p_t)[i] - \beta(p_d)[i]|
\end{equation}

\begin{figure*}[htb]
  \centering
  \includegraphics[width=0.95\linewidth]{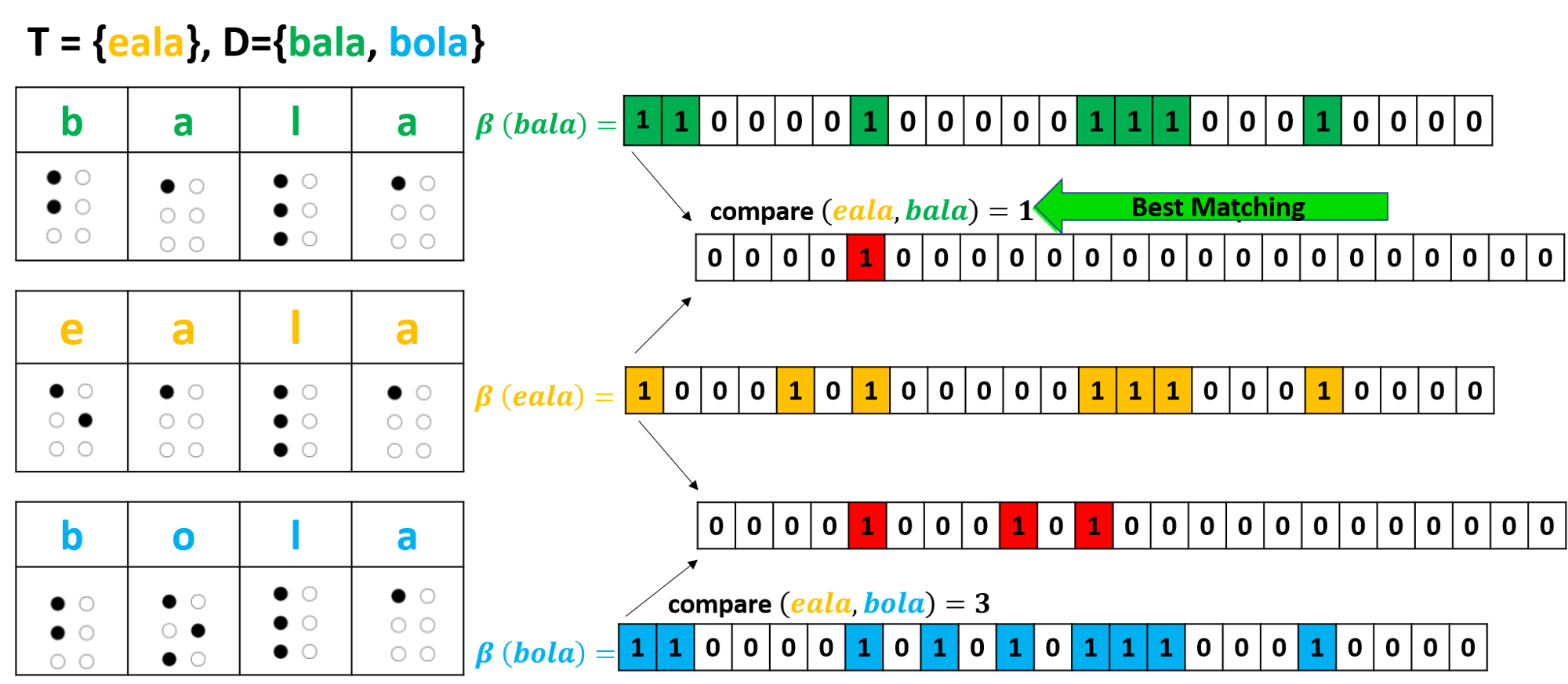}
  \caption{Example of the proposed algorithm's execution, with |T| = 1 and |D| = 2.}
  \label{fig:exemplo}
\end{figure*}

To minimize the Equation \ref{eq:compare}, the algorithm calculates the best match for each word $ p_t \in T $. Then we check for each word $p_t \in T$, the value corresponding to all words $ p_d \in D $, with the same lenght. If $ p_d $ minimizes this equation and is equals $ p_t $, then we consider $p_t$ grammatically true, otherwise we replace the referred word. Algorithm \ref{algoritmo} details the calculus of the $compare(p_t, p_d)$ function, and \autoref{fig:exemplo} presents a graphical example of using the algorithm with | T | = 1 and | D | = 2.


\begin{algorithm}
\caption{Revision on the trascription result.}
\label{alg:SSadapted}
\begin{algorithmic}[1]
\REQUIRE {$T, D$} 
  \FOR {i = 0 to T.size()}
      \STATE min = $\infty$
      \STATE aux = 0
      \FOR {j = 0 to D.size()}
            \IF {(T[i].size() == D[j].size()}
                \STATE diff = compare (T[i],D[j])
                \IF {(diff < min)}
                    \STATE min = diff
                    \STATE aux = j
                \ENDIF
            \ENDIF
       \ENDFOR
       T[i] = D[j]
  \ENDFOR
  \RETURN T
\end{algorithmic}
\label{algoritmo}
\end{algorithm}

\section{Experimental evaluation}

We evaluate our revision proposed method's robustness in two sets of experiments. In both, we imposed errors and submitted them to our proposal. In experiment A, we randomly added different percentages of error in each word, working directly with the text, so we do not recognize the braille character in images. In experiment B, we executed the full pipeline on synthetic images of braille texts, i.e., as described in Section \ref{sec:approach}. We adopted synthetic images in detriment to real ones to give more control to the recognition process and consequently have more reliable results on the revision proposal.

We compared our results with those obtained by the pySpell-Checker library \cite{pyspellcheker}. We contrasted our results with this library because it uses the Levenshtein distance, a metric commonly used by spellcheckers \cite{miller2009}. Besides using the Levenshtein Distance, the pySpellChecker method relies on a word frequency list (a dictionary with word frequencies). Higher frequency values represent the most frequently used words, so they are more likely to be the correct results. Another feature of pySpellChecker is that if a searched word is not in the dictionary, it returns a suggestion, if and only if a cost function is lower than a threshold. Otherwise, this library does not suggest any word as a correction.

We detail the experimental database in Subsection \ref{sub:database}. In Section \ref{sub:metrics}, we present the evaluation metrics we use on the experiments. Sections \ref{sub:experimentA} and \ref{sub:experimentB} present the experimental setup as well as the results and discussions concerning the experiments A and B. 

\subsection{Dataset}
\label{sub:database}


We carried out experiments based on ten texts and one dictionary. We got these texts from classic literature books of public domain\footnote{Books collected from the website: \url{http://www.dominiopublico.gov.br/}}. \autoref{tab-books} shows details of the books, with name and author. We select only the first chapter (poem, scene, or act) of each book in the experiments.

\begin{table}[htb]
\caption{Selected books and respective authors, employed to build our experimental dataset.}
  \label{table-parametros}
\begin{tabular}{lc}
\toprule
\textbf{Book} & \textbf{Author}\\
\midrule
Dom Casmuro & Machado de Assis\\
Iracema & José de Alencar \\
Poemas de Álvaro de Campos & Fernando Pessoa \\
Quincas Borba & Machado de Assis \\
A Morte do Lidador & Alexandre Herculado \\
Macbeth & William Shakespeare \\
A Dívida & Artur Azevedo \\
Alma Inquieta & Olavao Bilac \\
A Dama do Pé-De-Cabra & Alexandre Herculano \\
A Igreja do Diabo & Machado de Assis \\
\bottomrule
\end{tabular}
\label{tab-books}
\end{table}




As for the dictionary, as we compare our method with the pySpell-checker library, we select the dictionary recommended on that library website, which has a link to the repository of the WordFrequency project\footnote{https://github.com/hermitdave/FrequencyWords}. This repository has dictionaries in different languages. We selected the dictionary in the Brazilian Portuguese language, which has 848,043 words accompanied by the frequency list.

We preprocessed the dictionary by removing words containing special characters not belonging to the Portuguese language, i.e., those different from accented letters (á, à, â, ã, é, ê, í, ó, ô, õ, ú, and ç), and the punctuation characters hyphen (-) and apostrophe (` or '). This processing step removed 12,400 words from the dictionary. We kept these characters because they are common in the writing words in the Portuguese language.  This chosen dictionary is the reference for the pySpellChecker library, and besides the words, it also presents their frequency list. In our trials, we consider the full version (i.e., words and frequencies) for pySpellChecker, and only the word list for our approach. The dataset we use in this paper is available on \url{http://link_ommited_due_revision_purposes}.

\subsection{Evaluation Metrics}
\label{sub:metrics}

To compare the results of the experimental instances, we adopted two metrics: the Levenshtein distance and the percentage of correctly transcribed words (or hit rate).

The Levenshtein distance (or edition distance) calculates the minimum number of operations (inserting, deleting, or replacing a character) required to transform one string into another. This is a commonly used metric in spellcheckers, as it is useful in determining how similar two strings are \cite{miller2009}. 

The hit rate is a straightforward metric that computes the number of words correctly transcribed/corrected concerning the ground-truth. How higher this value, the better the experimental results. With the Levenshtein's distance, lower values are better. In unlikely to get all the words right, so lower values for the Levenshtein distance show that the words incorrectly transcribed are more similar to the reference set.

In addition, we employed three measures to understand the behavior of the proposed method in the experiments: char recognition error, word recognition error, and words found in the dictionary. The char recognition error relates to the recognition algorithm.  By the same logic, the word recognition error shows the words correctly detected. These metrics allow analyzing quantitatively the impact of each type of noise used in experiment B.  The percent of words in the dictionary represents the number of recognizable words by the dictionary. This metric plays important information by denoting the upper-bound values for each experiment.

\subsection{Experiment A: random error}
\label{sub:experimentA}

For each text on the dataset, we create a set $T$ with all its words. Then we convert each word $p_x \in T$ to a six-bin-code representation, using the $\beta$ function. Later, we impose an error on this code vector, flipping the values of $\beta(p_x)$ randomly.  For instance, considering word ``computação'' (with length 10), and $|\beta ($``computação''$)| = 60$, if we add 10\% error, it means that the values of 6 positions (10\% of 60) will be changed. 

After adding the error in each word $p_x \in T$, we convert the \textit{corrupted} vector back to a word using the $\psi$ function. Finally, for each word, we executed a correction proposal algorithm on the induced grammatical error.

We test tweve imposed error instances (from 2.5 to 30\% with a step of 2.5\%) in every text on the dataset. With these values, it was possible to assess the proposed method with higher error rates, which were more significant than the errors caused by the noise added in experiment B (Subsection \ref{sub:experimentB}) that simulate real braille transcription problems. \autoref{tab-results-A} present the results of experiment A. The first column shows the percentage of errors added in each word. The second group of columns shows the average Levenshtein distance. And the last group of columns presents the hit rate.

\begin{table}[htb]
\caption{Results of experiment inducing random error to the texts. We compare our method (ours) with pySpellChecker \cite{pyspellcheker} library on the Levenshtein distance and hit words percentage metrics. Best values on each metric in \textbf{bold}}
\label{tab-results-A}
\begin{tabular}{c|cc|cc}
\toprule
 \textbf{\% of error}& \multicolumn{2}{c|}{\textbf{Levenshtein dist.}} & \multicolumn{2}{c}{\textbf{\% of hit}} \\
  \textbf{added} & \textbf{ours} & \textbf{pySpell} & \textbf{ours} & \textbf{pySpell}\\
\midrule
2.5     & \textbf{0.10}          & 0.12 & \textbf{96.3}     & 89.9 \\
5.0     & \textbf{0.21} & 0.77         & \textbf{86.3}     & 52.0\\
7.5     & \textbf{0.44} & 1.45         & \textbf{70.5}     & 30.1 \\
10.0    & \textbf{0.87} & 1.98          & \textbf{45.9}     & 13.9 \\
12.5    & \textbf{1.25} & 2.39          & \textbf{34.0}     & 12.3 \\
15.0    & \textbf{1.62} & 2.63          & \textbf{26.9}     & 12.2  \\
17.5    & \textbf{2.16} & 3.10          & \textbf{10.0}      & 0.6  \\
20.0    & \textbf{2.43} & 3.25         & \textbf{6.4}      & 0.7 \\
22.5    & \textbf{2.73} & 3.44          & \textbf{4.0}      & 0.2  \\
25.0    & \textbf{2.96} & 3.60          & \textbf{2.6}      & 0.4 \\
27.5    & \textbf{3.14} & 3.70          & \textbf{2.1}      & 0.3 \\
30.0    & \textbf{3.35} & 3.81          & \textbf{1.4}      & 0.2 \\
\bottomrule
\end{tabular}
\end{table}

\begin{figure}[ht]
  \centering
  \includegraphics[width=1\linewidth]{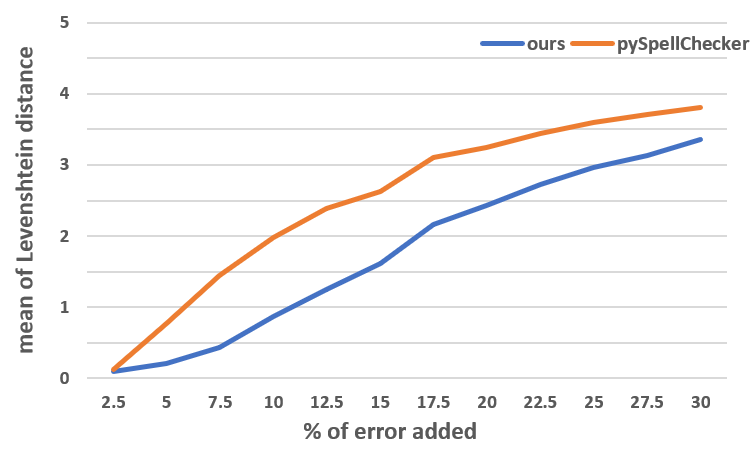}
  \caption{Results of experiment inducing random error to the texts. We compare our method (ours) with pySpellChecker \cite{pyspellcheker} library on the Levenshtein distance. Higher values are better.Our proposal outperforming the other competitor in every tested situation.}
  \label{fig:resultLevenshtein}
\end{figure}

\begin{figure}[ht]
  \centering
  \includegraphics[width=1\linewidth]{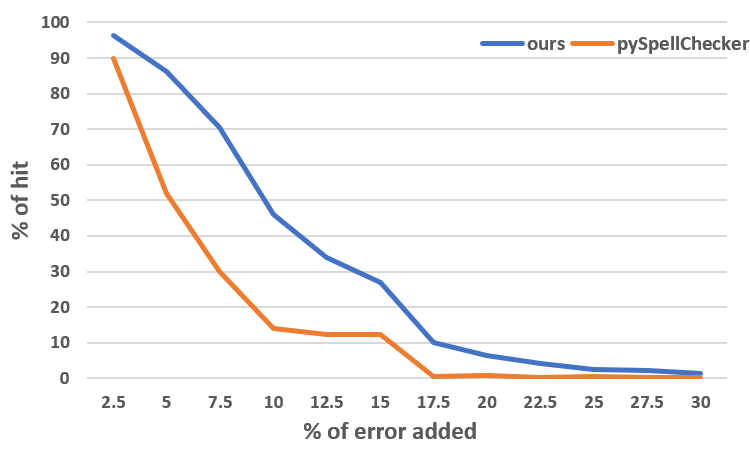}
  \caption{Results of experiment inducing random error to the texts. We compare our method (ours) with pySpellChecker \cite{pyspellcheker} library on the hit words percentage. Lower values are better. Our proposal outperforming the other competitor in every tested situation.}
  \label{fig:resultHit}
\end{figure}

Figures \ref{fig:resultLevenshtein} and \ref{fig:resultHit} presents the results of \autoref{tab-results-A} as graphs. The x-axis of the graphs refers to the percentage of error added in each experimental instance. The y-axis shows the results of each of the metrics. \autoref{fig:resultLevenshtein} represents the average Levenshtein distance for each word of the texts, and \autoref{fig:resultHit} represents the percentage of the hit.

\subsection{Experiment B: noise in synthetic image}
\label{sub:experimentB}

\begin{figure*}[htb]
  \centering
  \includegraphics[width=1\linewidth]{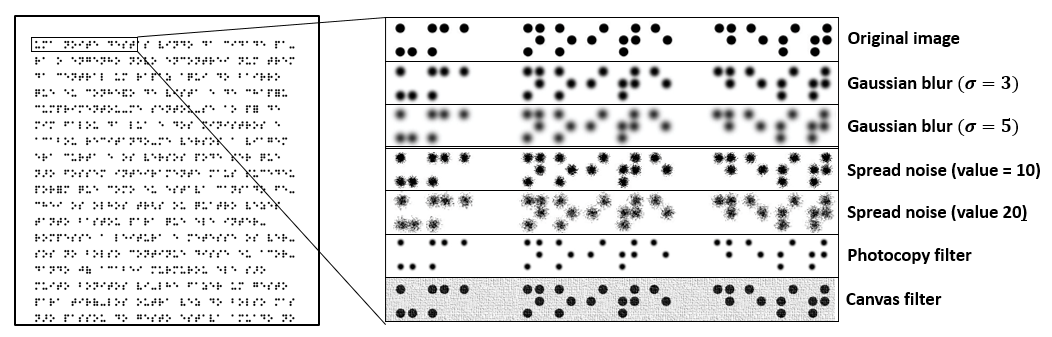}
  \caption{Experimental dataset. We converted the original braille texts into the image representation. Then we applied some image processing filters to induce errors in the recognition algorithm. We present each of these filters with the respective name/parameters.}
  \label{fig:test-data}
\end{figure*}

To synthetically produce braille documents from the original texts, we first performed a preprocessing step similar to the previous experiment. Besides the char removal, we also got rid of digits and uppercase letters. This last preprocessing step guarantees non-extra-braille characters addition because, in the braille system, number and capital letters representation demand a previous extra char.

We converted the preprocessed texts to braille with the Braille Fácil software \cite{braille-facil}. We converted the output braille texts to plain white images with black dots. We also imposed some transformations on the pictures to simulate some variability on the data and test our recognition algorithm. We carried out six image transformations on the Gimp software \cite{gimp}: in the first two, we applied Gaussian blur filters with sigma values of 3.0 and 5.0; in the other two, we randomly spread the braille dots with a variation value of 10 and 20 pixels on both vertical and horizontal direction; the last two correspond to the combination of several processing techniques, producing the artistic filters named photocopy and canvas. In \autoref{fig:test-data}, we present the transformation we made on the images applied to a braille document segment.

The error rates in the transcription process of experiment B are in \autoref{tab:errorrates} - each line refers to a type of noise added to the images. The last two columns represent the percentage of error in chars and word recognition, respectively.

\begin{table}[htb]
\caption{Recognition of braille characters/words on the proposed dataset. Column chars refers to the recognition rate of our recognition stage. Columns words refer to the word ratio returned as true on the dictionary}
\label{tab:errorrates}
\begin{tabular}{lcc}
\toprule
 & \multicolumn{2}{c}{\textbf{Recognition error}} \\ \cline{2-3} & \textbf{chars} & \textbf{words} \\ \midrule
Plain                           & 100.0 & 97.4 \\
Gaussian blur ($\sigma = 3.0$)  & 96.6 & 92.6 \\
Gaussian blur ($\sigma = 5.0$)  & 95.7 & 91.7 \\
Spread noise ($10$)             & 96.3 & 92.1 \\
Spread noise ($20$)             & 89.5 & 83.8 \\
Photocopy filter                & 90.7 & 90.5 \\ 
Canvas filter                   & 97.4 & 87.6 \\ 
\bottomrule
\end{tabular}
\end{table}

\autoref{tab-results-B} presents the results of experiment B. The first column describes the type of noise used. The second column group shows the Levenshtein Distance for each analyzed method. Finally, the last group of columns shows the hit percentage for each technique. In the last line of this table, we present the average of the results. 

\begin{table}[htb]
\caption{Results of experiment inducing noise to the document images. We compare our method (ours) with pySpellChecker \cite{pyspellcheker} library on the Levenshtein distance and hit words percentage metrics. Best values on each metric in \textbf{bold}}
  \label{tab-results-B}
\begin{tabular}{l|cc|cc}
\toprule
                    &  \multicolumn{2}{c|}{\textbf{Levenshtein dis.}}& \multicolumn{2}{c}{\textbf{\% of hit}}\\
  \textbf{noise}    &   \textbf{ours} & \textbf{pySpell} &  \textbf{ours} & \textbf{pySpell}\\
\midrule
Gaussian blur ($\sigma = 3.0$)   & 0.12&\textbf{0.08}& \textbf{94.8}  & 94.1 \\ 
Gaussian blur ($\sigma = 5.0$)  &0.13&\textbf{0.08}& \textbf{94.3}     & 93.8\\
Spread noise (10)               &0.17&\textbf{0.13}& \textbf{93.9}     & 93.6\\
Spread noise (20)  &0.24 &\textbf{0.19}& \textbf{87.8}              & 86.8\\ 
Photocopy filter         &0.19&\textbf{0.12}  & 91.2              & \textbf{91.5}\\ 
Canvas filter               &\textbf{0.37}&0.39 & \textbf{89.0}    & 88.6 \\
\midrule
Average results       &0.20&\textbf{0.17}& \textbf{91.8}     & 91.4\\
\bottomrule
\end{tabular}
\end{table}

\begin{figure}[htb]
  \includegraphics[width=1\linewidth]{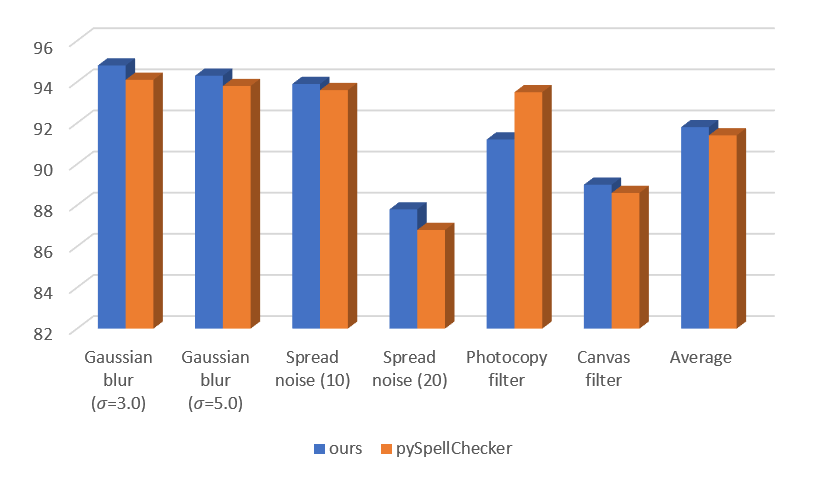}
  \caption{Results of experiment inducing noise to the document images. We compare our method (ours) with pySpellChecker \cite{pyspellcheker} library on the hit words percentage. Higher is better.}
  \label{fig:graficoHit}
\end{figure}

\begin{figure}[htb]
  \includegraphics[width=1\linewidth]{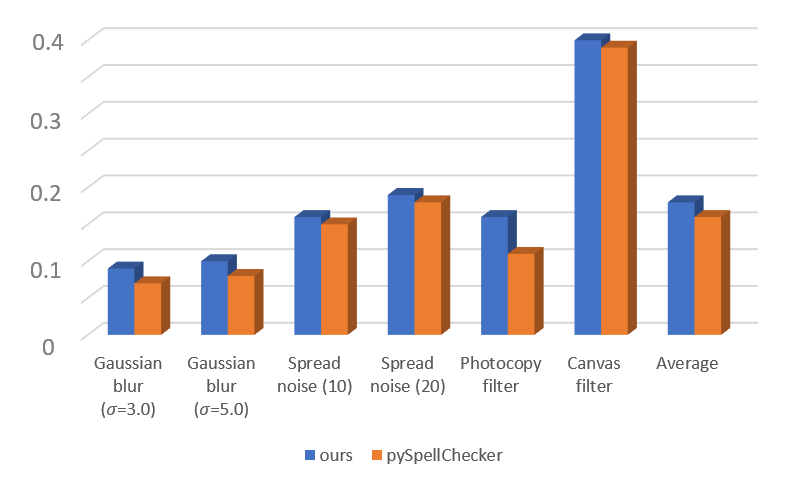}
  \caption{Results of experiment inducing noise to the document images. We compare our method (ours) with pySpellChecker \cite{pyspellcheker} library on Levenshtein distance. Lower is better.}
  \label{fig:graficoLevenshtein}
\end{figure}

Figures \ref{fig:graficoLevenshtein} and \ref{fig:graficoHit} present the results of the \autoref{tab-results-B} graphically.  In \autoref{fig:graficoLevenshtein} we compare the Levenshtein Distance, and in \autoref{fig:graficoHit} we compare the percentage of hit. 

\subsection{Discussion}
\label{sub:analysis}
In experiment A, the proposed algorithm outperformed the pySpell-Cheker library for every tested instances, considering the two metrics. These results show our approach's robustness when we found a higher transcription error level, i.e., the higher the percentage of induced error, the results from both methods get worse. Still, proportionally our approach shows far better results than pySpellChecker.

Concerning experiment B, we also outperform pySpellChecker considering the \% hit metric (except for the Photocopy Filter noise). Regarding Levenshtein's distance, our method presents higher values.  The pySpellChecker algorithm minimizes the Levenshtein's distance, and we expected lower values on such metrics. This behavior did not occur in experiment A because our method's accuracy rate was proportionally higher, affecting the Levenshtein distance results.

The data presented in \autoref{tab:errorrates} help to understand the behavior of the method in the experiments. This table shows the percent correctly recognized chars and the percentage of words identified with each type of induced noise related to the ground truth that also exists in the dictionary. This percentage of words represents the maximum accuracy that our method can achieve in experiment B. It is impossible to accurately suggest a word if it does not exist in the dictionary. In this context, we found that the more representative the dictionary is about the transcribed text, the better our method's results should be.


The information regarding the character recognition error (shown in \autoref{tab:errorrates} is essential for understanding the experimental results. The more unsuccessfully recognized characters we have, the more challenging the text review process is.

We should note that the method of reviewing transcribed braille texts proposed in this work compares only words of the same size. Given an entry of size $n$, our approach only compares terms of length $n$. This is a limitation of classical methods based on Levenshtein's distance, which we plan to improve in future works.

Despite this limitation, our results outperformed by far on most experimental situations imposed, which confirms that our method is effective and can offer a reliable tool for reviewing texts transcribed from Braille to Portuguese.

\section{Conclusions and Future works}

In this paper, we presented a full pipeline to transcribe and revise braille texts in images. To the best of our knowledge, our proposal is the first braille-cell-oriented approach to correct misrecognized words in texts. Results show that our method outperforms standard dictionary-based approaches. This paper represents a study regarding the revision process, and in further studies, we aim to test our proposal in real-braille-images. However, these initial results show a remarkable potentiality of our approach as a tool for image-based-braille-transcription systems.

Future directions of this seminal work include to test our proposal on real braille-printed images, single and double-sided. We also aim to implement a functional mobile application and evaluate its human-interaction aspects following works like \cite{farinella2015mobile}. Another essential trend nowadays in computer vision is deep learning, and proposals like U-Net \cite{ronneberger2015u} and BraUNet \cite{li2020optical} must be considered to improve the recognition stage of our pipeline.  We also aim to work with other dictionaries and extend our proposal to deal with different lengths of resulting words. 

We focus our revision method on correct mis-recognized words that are expected to be orthographically correct on the original document. However, when dealing with real human-produced texts, some typos and grammatical errors may occur. Our work can also act in this domain, being an orthographic and grammatical revisor. Finally, we aim to apply natural language processing algorithms to, besides an orthographic suggestion, also consider semantic features to improve our revision stage.



\bibliographystyle{ACM-Reference-Format}
\bibliography{sample-base.bib}

\end{document}